\documentclass[11pt,a4paper]{article}
\pdfoutput=1
\usepackage{jheppub}
\usepackage[T1]{fontenc}
\usepackage{lmodern}
\usepackage{microtype}
\usepackage{booktabs}
\usepackage{tabularx}
\usepackage{array}
\usepackage{bm}
\usepackage{enumitem}
\usepackage{float}
\usepackage{cleveref}

\makeatletter
\def\csid@tablecaptype{table}
\long\def\@makecaption#1#2{%
  \ifx\@captype\csid@tablecaptype\else\vskip\abovecaptionskip\fi
  \sbox\@tempboxa{\small #1. #2}%
  \ifdim \wd\@tempboxa >\hsize
    \small #1. #2\par
  \else
    \global \@minipagefalse
    \hb@xt@\hsize{\hfil\box\@tempboxa\hfil}%
  \fi
  \ifx\@captype\csid@tablecaptype\vskip\abovecaptionskip\else\vskip\belowcaptionskip\fi}
\makeatother

\newcommand{\supp}{\operatorname{supp}}
\newcommand{\argmin}{\operatorname*{arg\,min}}
\newcommand{\Var}{\operatorname{Var}}
\newcommand{\Cov}{\operatorname{Cov}}
\newcommand{\CCC}{\operatorname{CCC}}
\newcommand{\logit}{\operatorname{logit}}

\title{Cardinality-Stratified Interaction Decomposition for Interpretable Pairwise and Higher-Order Structure in Transactional Basket Data}

\author{Hidetoshi Kawase}
\author{and Toshihiro Ota}
\affiliation{CyberAgent, Inc., Shibuya, Tokyo 150--6121 Japan}
\emailAdd{kawase\_hidetoshi@cyberagent.co.jp}
\emailAdd{ota\_toshihiro@cyberagent.co.jp}

\abstract{
Transactional basket data can reveal associations among items, but observed co-occurrence conflates item-specific relations with basket-size structure and unmodeled higher-order dependence.
We introduce Cardinality-Stratified Interaction Decomposition (CSID), an interpretable framework that decomposes log-odds contrasts stratified by the number of remaining items into item-set-specific and cardinality-common components, without fitting a global joint distribution.
CSID uses an information-weighted, gauge-constrained ridge projection to estimate pair and triple components and to diagnose higher-order contributions to pairwise structure.
CSID is designed primarily for interpretable decomposition of association structure rather than for full-distribution prediction.
In a simulation with zero pair effects, increasingly strong small-basket cardinality potentials drive ordinary Ising couplings spuriously negative, whereas CSID pair estimates remain centered near zero.
Detection power rises with the magnitude of planted triple effects, and local deprojection reduces pair-coefficient RMSE from 0.244 to 0.073.
Across three grocery datasets, high-information triple components are reproducible over time.
In the matched cross-period partial-transfer evaluation, transferred CSID triple components show closer agreement with later-period stratified contrasts than the nodewise-symmetrized cardinality-aware higher-order pseudolikelihood comparator, with gains in weighted Lin's concordance correlation of 0.038--0.122.
These results support CSID as an exploratory and interpretable decomposition framework for pairwise and higher-order association structure in transactional data.
}

\makeatletter
\gdef\@fpheader{ \vspace{1em} }
\makeatother

\begin{document}
\maketitle
\flushbottom

\section{Introduction}

Transactional basket data record the set of items selected on a single purchase occasion.
A central objective of basket analysis is to characterize how items relate to one another: which items tend to reinforce each other's purchase, which items compete for inclusion in the same basket, and how these relations change in the presence of additional items.
Classical market-basket analysis identifies frequent itemsets and association rules using measures such as support, confidence, and lift \cite{agrawal1993mining}.
These measures are effective for discovering co-occurrence patterns, but they do not directly provide a signed conditional relation between items.
An observed co-occurrence may simultaneously reflect item popularity, basket-size composition, lower-order associations, and genuinely higher-order interaction.

A binary Ising model, or equivalently a pairwise binary log-linear model, provides a natural representation for this purpose.
Its field $h_i$ describes the baseline propensity to purchase item $i$, whereas its coupling $J_{ij}$ contributes to the conditional association between items $i$ and $j$ after the remaining items are fixed.
Positive and negative couplings can therefore be interpreted as complement-like and exclusion-like purchase tendencies within a common signed network.
Such a representation is potentially useful for cross-selling, competition analysis, and assortment diagnosis, and inverse-Ising models have previously been applied to market-basket data \cite{valle2019market}.

The interpretation of $J_{ij}$ becomes difficult, however, when the basket-size distribution is not represented separately.
Retail transaction data often contain many single-item and small baskets.
The tendency not to add a second or subsequent item is a basket-wide property that does not depend on a particular pair of item labels.
An ordinary pairwise Ising model has no explicit component for this global cardinality structure and may reproduce small baskets by shifting many pairwise couplings in the negative direction.
A negative $J_{ij}$ then becomes ambiguous: it may represent an item-specific exclusionary relation, or it may merely absorb common small-basket pressure.

We address this problem by augmenting the binary log-linear reference model with a cardinality potential $\alpha_{|x|}$, which depends only on the number of items in the basket.
The cardinality potential represents item-label-invariant tendencies toward single-item, small, or large baskets, whereas the pairwise coefficients represent deviations associated with particular item pairs.
The purpose of this separation is not merely to improve predictive fit, but to define the sign of an item interaction relative to a common basket-cardinality environment.

A second interpretive difficulty arises when the basket structure is not pairwise additive.
For example, a positive milk--cereal association may weaken in baskets that also contain fruit, or a snack--dip association may saturate when crackers are present.
Such changes in a pair association induced by a third item are third-order interactions.
If they are omitted, their contributions are absorbed into pairwise projections, so an estimated pair coefficient may combine a pair-specific component with the effects of surrounding items.
Recovering an interpretable pair structure therefore requires both the separation of global cardinality pressure and the diagnosis of higher-order contributions.

This paper introduces Cardinality-Stratified Interaction Decomposition (CSID), a framework for estimating pairwise and higher-order interaction components from sparse transactional data.
For a focal item set $T$, we begin with the corresponding higher-order log-odds contrast at a fixed configuration of the remaining items.
Under a cardinality-aware higher-order binary log-linear reference model, this contrast separates into interactions containing $T$ and a finite difference of the cardinality potential.
Because conditioning on the full remaining-item configuration produces extremely sparse cells, CSID instead stratifies by the number of purchased items outside $T$ and fits the additive projection
\begin{equation*}
\Lambda_{T,r}\approx\theta_T+\beta_{k,r}.
\end{equation*}
Here $\theta_T$ is an item-set-specific projection component and $\beta_{k,r}$ is shared by interactions of order $k$ in rest-cardinality stratum $r$.
The projection is estimated by an information-weighted, gauge-constrained ridge criterion and does not require evaluation of a global partition function.
We use CSID-$k$ for the order-$k$ estimator, in particular CSID-2 for pairs and CSID-3 for triples.

The primary objective of CSID is interpretive rather than predictive.
It is designed to separate an item-set-specific association component from a rest-cardinality-common environment and to diagnose how higher-order structure alters lower-order projections.
In the empirical evaluation, we use a cardinality-aware higher-order pseudolikelihood model (CA-HOPL)---a conditional model with pair, triple, and flexible rest-cardinality terms---as an external comparator.
This comparison assesses the reproducibility and transferability of the contrast components targeted by CSID rather than universal predictive superiority.

The contributions of this work are as follows.

\begin{enumerate}[leftmargin=2.2em]

\item
We formulate a cardinality-aware higher-order binary log-linear reference model and derive the decomposition of a fixed-rest interaction contrast into item-set interaction terms and a finite difference of the cardinality potential.
This decomposition explains how global small-basket pressure can contaminate the signs of ordinary pairwise Ising couplings.

\item
We define CSID-$k$ as a gauge-constrained, information-weighted ridge projection of rest-cardinality-stratified interaction contrasts.
We derive closed-form alternating updates and establish uniqueness of the gauge-fixed item-set block and the informative rest-cardinality strata.

\item
We characterize how third-order interactions enter pairwise projections and construct a local reweighting procedure designed to adjust pairwise projections for estimated third-order contributions before the pairwise projection is re-estimated.

\item
We evaluate the framework through simulations of cardinality-induced contamination of pair estimates, randomized sparse third-order effects, and third-order saturation, followed by analyses of three public grocery datasets.
The empirical evaluation examines cross-period stability, continuous transfer of triple-specific effects, conditional-odds interpretation, pairwise deprojection, and robustness to expansion of the surrounding item universe.

\end{enumerate}

\section{Related Work}

Classical market-basket mining searches for frequent itemsets and implication-style rules, while subsequent dependence-rule methods explicitly distinguish association from statistical dependence \cite{agrawal1993mining,brin1997beyond}, and systematic comparisons show that different objective interestingness measures rank the same patterns differently \cite{tan2004selecting,geng2006interestingness}.
Log-linear approaches instead represent multi-item associations through parameters of a contingency-table model and can screen interactions that are not explained by lower-order terms \cite{wu2003screening}.
CSID shares this emphasis on interpretable interaction contrasts but addresses sparse high-dimensional tables by aggregating over rest-item identities within cardinality strata.

Hierarchical log-linear models provide the standard interaction parameterization for multiway contingency tables \cite{bishop1975discrete}.
Marginal log-linear models further distinguish interactions defined in different margins and clarify the consequences of marginalization \cite{glonek1995multivariate,bergsma2002marginal}.
CSID builds on these established marginal interaction contrasts rather than proposing a new complete parameterization.
Its contribution is the information-weighted decomposition of these contrasts into item-set-specific and shared rest-cardinality components, together with gauge fixing and higher-to-lower-order diagnostics for sparse transactional data.

Pairwise maximum-entropy and inverse-Ising models offer signed conditional dependence parameters, with pseudolikelihood and related approximations enabling estimation in larger systems \cite{besag1975statistical,ravikumar2010high,aurell2012inverse,nguyen2017inverse}.
Whether higher-order terms are needed at all has been examined in detail in neural population data, where pairwise models already capture strongly collective states and a sparse network of low-order interactions improves the description further \cite{schneidman2006weak,ganmor2011sparse}.
Higher-order interactions are now studied systematically across complex systems \cite{battiston2020networks,battiston2021physics,matias2026statistical}, including statistically validated groups of co-purchased products in retail baskets \cite{betti2026identifying}.
Inverse-Ising methodology has been applied directly to transactional baskets \cite{valle2019market}, while multivariate logit models have represented sparse cross-category and selected higher-order effects in consumer choice \cite{hruschka2026analyzing}.
In marketing science, cross-category dependence in basket composition has long been represented through multivariate purchase-incidence and choice models \cite{manchanda1999shopping,russell2000analysis}.
Higher-order inverse problems have also been studied beyond pairwise Ising structure \cite{beentjes2020higher,decelle2025inferring}, and recent work addresses model selection and generative modeling with interactions of arbitrary order \cite{demulatier2020bayesian,declercq2026modeling}.
These approaches primarily estimate a joint or conditional model; CSID instead targets local pair and triple projection components without evaluating a global partition function.
To distinguish this estimand difference from the mere inclusion of higher-order and cardinality terms, our empirical evaluation directly compares CSID-3 with a cardinality-aware order-three pseudolikelihood model by testing how effects estimated in one period improve reconstruction of stratified contrasts in a later period.

Cardinality potentials are high-order factors whose value depends only on the number of active binary variables \cite{gupta2007efficient,tarlow2012fast,swersky2012cardinality}.
CSID uses a cardinality-aware model as an interpretive reference, but does not fit its global normalizing constant.
Its distinctive step is to stratify focal contrasts by rest cardinality, estimate the item-set-specific deviation from the common stratum environment, and use higher-order estimates to diagnose contributions absorbed into lower-order projections.

\section{Cardinality-Stratified Interaction Decomposition}

\subsection{Reference model and gauge freedom}

Let $V=\{1,\ldots,N\}$ be the item set and let
\begin{equation}
x=(x_1,\ldots,x_N)\in\{0,1\}^N
\end{equation}
represent one basket, with $x_i=1$ if item $i$ is present.
The basket cardinality is $|x|=\sum_i x_i$.

The ordinary $0/1$ pairwise Ising model is
\begin{equation}
P_{\mathrm{Ising}}(x)\propto
\exp\left\{
\sum_i h_i x_i+\sum_{i<j}J_{ij}x_i x_j
\right\}.
\end{equation}
To represent item-label-invariant basket-size structure, we first consider the cardinality-aware pairwise model
\begin{equation}
P(x)\propto
\exp\left\{
\alpha_{|x|}+\sum_i h_i x_i+\sum_{i<j}J_{ij}x_i x_j
\right\}.
\end{equation}
Here $\alpha_{|x|}$ captures single-item, small-basket, and large-basket tendencies, while $J_{ij}$ represents a pair-specific deviation from this common structure.

For a higher-order formulation, define $x_U=\prod_{i\in U}x_i$ for $U\subseteq V$ and consider the cardinality-aware log-linear model of maximum order $K$,
\begin{equation}
P(x)\propto
\exp\left\{
\alpha_{|x|}
+
\sum_{\substack{\varnothing\neq U\subseteq V\\ |U|\le K}}
\theta_U x_U
\right\}.
\label{eq:model}
\end{equation}
The singleton parameter is $\theta_{\{i\}}=h_i$, the pair parameter is $\theta_{\{i,j\}}=J_{ij}$, and $\theta_U$ for $|U|\ge3$ is a higher-order interaction.
The case $K=2$ is the cardinality-aware Ising model above.

For every order $k$,
\begin{equation}
\sum_{|S|=k}x_S=\binom{|x|}{k}.
\end{equation}
When all order-$k$ subsets are included in \cref{eq:model}, the transformation
\begin{equation}
\theta_S\mapsto\theta_S+c\quad(|S|=k),
\qquad
\alpha_m\mapsto\alpha_m-c\binom{m}{k}
\end{equation}
leaves the exponent, and hence the distribution, unchanged.
This is the structural gauge freedom between the order-$k$ interaction block and the cardinality potential.
When only a restricted family of order-$k$ subsets is included, however, their presence indicators need not sum to a function of $|x|$ alone, so this transformation does not generally preserve the distribution.

\subsection{Higher-order log-odds contrasts}

Let $T\subseteq V$ with $|T|=k$.
Write $y=x_{-T}$ for the configuration outside $T$, and let $r=|y|$.
For $A\subseteq T$, write $x_T=\mathbf 1_A$ for the state in which exactly the items in $A$ are present within $T$.
We define the order-$k$ log-odds contrast at fixed $y$ as
\begin{equation}
\Lambda_T(y)=
\sum_{A\subseteq T}(-1)^{k-|A|}
\log P(x_T=\mathbf 1_A,x_{-T}=y).
\label{eq:fixedcontrast}
\end{equation}

Substituting \cref{eq:model} into \cref{eq:fixedcontrast} gives the decomposition directly.
The normalizing constant cancels because
\begin{equation}
\sum_{A\subseteq T}(-1)^{k-|A|}=0.
\end{equation}
For an interaction monomial $x_U$, only interactions whose index set contains all elements of $T$ survive the contrast.
Writing each surviving index set uniquely as $U=T\cup W$, with $W\subseteq\supp(y)$, the total interaction contribution to the contrast is
\begin{equation}
\sum_{\substack{
W\subseteq\supp(y)\\
|T|+|W|\le K
}}
\theta_{T\cup W},
\end{equation}
where
\begin{equation}
\supp(y)
=
\{i\in V\setminus T:y_i=1\}.
\end{equation}
The cardinality term contributes
\begin{equation}
\sum_{A\subseteq T}
(-1)^{k-|A|}\alpha_{|y|+|A|}
=
\sum_{\ell=0}^{k}
(-1)^{k-\ell}\binom{k}{\ell}\alpha_{|y|+\ell}.
\end{equation}
Therefore,
\begin{equation}
\Lambda_T(y)=
\sum_{\substack{W\subseteq\supp(y)\\ |T|+|W|\le K}}
\theta_{T\cup W}
+
\Delta^k\alpha_{|y|},
\label{eq:generaldecomp}
\end{equation}
where
\begin{equation}
\Delta^k\alpha_r=
\sum_{\ell=0}^{k}(-1)^{k-\ell}
\binom{k}{\ell}\alpha_{r+\ell}.
\label{eq:finite_difference}
\end{equation}
In particular, when the maximum interaction order is $K=k$,
\begin{equation}
\Lambda_T(y)=\theta_T+\Delta^k\alpha_{|y|}.
\label{eq:exactdecomp}
\end{equation}

For $k=2$, write $p_{ab}(y)=P(x_i=a,x_j=b,y)$.
Then $\Lambda_{ij}(y)$ is the usual conditional pair log-odds ratio:
\begin{equation}
\begin{aligned}
\Lambda_{ij}(y)
&=
\log\frac{p_{11}(y)p_{00}(y)}
{p_{10}(y)p_{01}(y)}
\\
&\equiv \log OR_{ij\mid y}.
\end{aligned}
\label{eq:pair_or}
\end{equation}

For $k=3$, $T=\{i,j,\ell\}$,
\begin{equation}
\Lambda_{ij\ell}(y)=
\log OR_{ij\mid x_\ell=1,y}
-
\log OR_{ij\mid x_\ell=0,y}.
\label{eq:triple_or}
\end{equation}
Thus, the sign of $\Lambda_{ij\ell}(y)$ indicates whether the conditional association between $i$ and $j$ is stronger or weaker when item $\ell$ is present.
When $K=3$, the coefficient $\theta_{ij\ell}$ is the item-set-specific component of this change after removing the common cardinality contribution $\Delta^3\alpha_{|y|}$.
A positive $\theta_{ij\ell}$ represents reinforcement beyond the common cardinality effect, whereas a negative value represents weakening or saturation.
If $K>3$, \cref{eq:generaldecomp} shows that the contrast also contains higher-order interactions involving all three focal items.

\subsection{Projection after stratification by rest cardinality}

Conditioning on every full rest vector $y$ rapidly produces sparse cells.
We therefore stratify only by
\begin{equation}
R_T=|x_{-T}|.
\end{equation}
For $A\subseteq T$, define
\begin{equation}
p_{T,r}^{A}=P(x_T=\mathbf 1_A,R_T=r)
\end{equation}
and
\begin{equation}
\Lambda_{T,r}=
\sum_{A\subseteq T}(-1)^{k-|A|}
\log p_{T,r}^{A}.
\label{eq:stratifiedcontrast}
\end{equation}
Marginalizing the identities of rest items generally produces
\begin{equation}
\Lambda_{T,r}=\theta_T+\beta_{k,r}+\rho_{T,r},
\label{eq:projection_residual}
\end{equation}
where $\beta_{k,r}$ is common to order $k$ and rest cardinality $r$, while $\rho_{T,r}$ contains mediation and mixing effects induced by marginalization.
We therefore target the additive projection
\begin{equation}
\Lambda_{T,r}\approx\theta_T+\beta_{k,r}.
\label{eq:projection}
\end{equation}
When rest configurations are fully resolved, \cref{eq:exactdecomp} expresses the contrast through the model interaction $\theta_T$ at fixed $y$.
Rest-cardinality stratification instead uses the additive summary \cref{eq:projection}, in which $\theta_T$ denotes the item-set-specific component relative to the common order-$k$ environment.

\paragraph{Population target.}
For a fixed finite candidate family $\mathcal T_k$, define the informative index set
\begin{equation}
\mathcal I_+
=
\{(T,r):p_{T,r}^{A}>0\ \text{for every }A\subseteq T\}.
\label{eq:informative_index}
\end{equation}
For $(T,r)\in\mathcal I_+$, let
\begin{equation}
q_{T,r}=
\left(
\sum_{A\subseteq T}\frac{1}{p_{T,r}^{A}}
\right)^{-1},
\label{eq:population_weight}
\end{equation}
and set $q_{T,r}=0$ outside $\mathcal I_+$.
The additive representation in \cref{eq:projection} is invariant under
\begin{equation}
\theta_T\mapsto\theta_T+c,
\qquad
\beta_{k,r}\mapsto\beta_{k,r}-c.
\end{equation}
This projection-level nonidentifiability is distinct from the structural gauge freedom in \cref{eq:model}: it holds even when $\mathcal T_k$ is restricted and cannot necessarily be absorbed into $\alpha_{|x|}$.
To identify the two components, let $\bar g_T^\star>0$ be fixed reference weights normalized to sum to one.
The population CSID component is then the weighted projection
\begin{equation}
(\theta^\star,\beta^\star)
=
\argmin_{\theta,\beta}
\frac12\sum_{(T,r)\in\mathcal I_+}
q_{T,r}
(\Lambda_{T,r}-\theta_T-\beta_{k,r})^2,
\qquad
\text{subject to }\sum_{T\in\mathcal T_k}\bar g_T^\star\theta_T=0.
\label{eq:population_projection}
\end{equation}
Thus the estimand is defined by the candidate family, population cell probabilities, and gauge convention; it need not equal a natural parameter of the reference joint model.

\paragraph{Cell contrasts and information weights}

Given a sample of $M$ baskets $x^{(1)},\ldots,x^{(M)}$, for candidate $T\in\mathcal T_k$, rest cardinality $r$, and pattern $A\subseteq T$, define
\begin{equation}
n_{T,r}^{A}=
\#\{n:x_T^{(n)}=\mathbf 1_A,\ R_T^{(n)}=r\}.
\label{eq:cellcount}
\end{equation}
With smoothing constant $\varepsilon>0$,
\begin{equation}
\widehat\Lambda_{T,r}=
\sum_{A\subseteq T}(-1)^{k-|A|}
\log(n_{T,r}^{A}+\varepsilon).
\label{eq:estimatedcontrast}
\end{equation}
For $k=3$, $T=\{i,j,\ell\}$,
\begin{equation}
\widehat\Lambda_{ij\ell,r}=
\log\frac{
(n_{111,r}+\varepsilon)(n_{100,r}+\varepsilon)
(n_{010,r}+\varepsilon)(n_{001,r}+\varepsilon)
}{
(n_{110,r}+\varepsilon)(n_{101,r}+\varepsilon)
(n_{011,r}+\varepsilon)(n_{000,r}+\varepsilon)
}.
\label{eq:triple_empirical}
\end{equation}
Equivalently,
\begin{equation}
\widehat\Lambda_{ij\ell,r}=
\log \widehat{OR}_{ij\mid x_\ell=1,R_T=r}-
\log \widehat{OR}_{ij\mid x_\ell=0,R_T=r}.
\label{eq:triple_or_empirical}
\end{equation}

Using the delta-method variance approximation for log cell counts \cite{agresti2013categorical}, define
\begin{equation}
w_{T,r}=
\left(
\sum_{A\subseteq T}\frac{1}{n_{T,r}^{A}+\varepsilon}
\right)^{-1},
\qquad
\mu_T=\sum_r w_{T,r}.
\label{eq:weights}
\end{equation}
Strata with sparse cell counts therefore receive low information weight.

\subsection{Gauge-constrained estimation and closed-form updates}

The finite-sample estimator uses the corresponding gauge constraint.
Let $g_T>0$ denote fixed reference weights used to select one representative from the equivalence class described above.
A natural default is $g_T=\mu_T$, using the information weights in \cref{eq:weights}.
We impose
\begin{equation}
\sum_{T\in\mathcal T_k}g_T\theta_T=0.
\label{eq:gauge}
\end{equation}
When $\mathcal T_k$ is restricted, \cref{eq:gauge} should be understood as a reference-weighted centering convention within the candidate family.
Hence, within the chosen candidate family, the sign of $\widehat\theta_{\{i,j\}}$ indicates whether the pair-specific projection is stronger or weaker than the reference-weighted average pair under the same rest-cardinality environment.

To estimate \cref{eq:projection}, we define CSID-$k$ as
\begin{equation}
(\widehat\theta,\widehat\beta)=
\argmin_{\theta,\beta}
\left\{
\frac12\sum_{T\in\mathcal T_k}\sum_r
w_{T,r}
(\widehat\Lambda_{T,r}-\theta_T-\beta_{k,r})^2
+
\frac{\lambda_k}{2}\sum_{T\in\mathcal T_k}\theta_T^2
\right\},
\label{eq:objective}
\end{equation}
subject to the gauge constraint \cref{eq:gauge}.
Dividing \cref{eq:objective} by $M$ makes its relation to \cref{eq:population_projection} explicit: $w_{T,r}/M\to q_{T,r}$ and $\widehat\Lambda_{T,r}\to\Lambda_{T,r}$.
Hence, for fixed $\varepsilon$ and a fixed candidate family, the empirical minimizer converges to the population projection provided that the gauge-constrained population projection is uniquely identified, the normalized gauge weights converge to $\bar g_T^\star$, and the ridge parameter $\lambda_{k,M}$ at sample size $M$ satisfies $\lambda_{k,M}/M\to0$ \cite{vandervaart1998asymptotic}.
For the default $g_T=\mu_T$, the limiting normalized gauge weight is $\bar g_T^\star=(\sum_rq_{T,r})/(\sum_{U\in\mathcal T_k}\sum_rq_{U,r})$.

For finite $M$, $\lambda_k>0$ ensures a unique gauge-fixed $\theta$ block and unique $\beta_{k,r}$ for strata with $\sum_Tw_{T,r}>0$.
Zero-information strata are fixed to zero by convention.
The resulting $\widehat\theta_T$ therefore depends on the candidate family, information weights, gauge, and ridge penalty.

Initialize $\beta_{k,r}^{(0)}=0$ and define at iteration $t$
\begin{equation}
S_T^{(t)}=
\sum_r w_{T,r}
(\widehat\Lambda_{T,r}-\beta_{k,r}^{(t)}).
\label{eq:residualsum}
\end{equation}
The Lagrange multiplier for the gauge constraint is
\begin{equation}
\gamma^{(t)}=
\frac{
\displaystyle\sum_{T\in\mathcal T_k}
\frac{g_T S_T^{(t)}}{\mu_T+\lambda_k}
}{
\displaystyle\sum_{T\in\mathcal T_k}
\frac{g_T^2}{\mu_T+\lambda_k}
},
\label{eq:gamma}
\end{equation}
and the exact update of the item-set block is
\begin{equation}
\theta_T^{(t+1)}=
\frac{S_T^{(t)}-\gamma^{(t)}g_T}
{\mu_T+\lambda_k}.
\label{eq:theta_update}
\end{equation}
The rest-cardinality-common block is then updated by
\begin{equation}
\beta_{k,r}^{(t+1)}=
\frac{
\displaystyle\sum_{T\in\mathcal T_k}w_{T,r}
(\widehat\Lambda_{T,r}-\theta_T^{(t+1)})
}{
\displaystyle\sum_{T\in\mathcal T_k}w_{T,r}
}.
\label{eq:beta_update}
\end{equation}
Unobserved $(T,r)$ cells are omitted by setting $w_{T,r}=0$.
Each step exactly minimizes one block, so the objective decreases monotonically and converges to the unique optimum for $\theta$ and the informative $\beta$ strata under the stated convention.

To prioritize candidate item sets for subsequent analysis, we normalize each estimated component by its approximate estimation scale.
Using the diagonal curvature $\lambda_k+\mu_T$ of \cref{eq:objective} as an information approximation gives
\begin{equation}
\widehat s_T
=(\lambda_k+\mu_T)^{-1/2},
\qquad
z_T=
\frac{\widehat\theta_T}
{\widehat s_T}.
\label{eq:zscore}
\end{equation}
Because this diagonal-curvature scale ignores uncertainty in the common effects, candidate overlap, and data-dependent screening, we use $|z_T|$ only as an information-normalized screening score, not for formal inference.

\paragraph{Computational cost.}
Exhaustive order-$k$ candidate generation is $O(N^k)$ and is intended only for moderately sized item universes.
For the pair-and-triple implementation used here, the counting pass over a fixed candidate family uses hash-indexed candidates and costs $O\{\sum_n[\binom{m_n}{2}+\binom{m_n}{3}]\}$ for basket sizes $m_n$.
The stratified arrays require $O((N+Q_2+Q_3)R)$ storage, where $Q_k=|\mathcal T_k|$ and $R$ is the number of retained rest-cardinality layers, and each alternating fitting iteration is $O(Q_kR)$.
For larger universes, the implementation constructs candidates from frequent-pair triangles and bounded item neighborhoods before exact counting, analogously to the candidate-generation step of the Apriori algorithm \cite{agrawal1994fast}.

\subsection{Pairwise deprojection of third-order effects}

Under \cref{eq:model} with $K=3$, applying \cref{eq:generaldecomp} to the pair $T=\{i,j\}$ gives
\begin{equation}
\Lambda_{ij}(y)=
J_{ij}
+
\sum_{\ell\in\supp(y)}\theta_{ij\ell}
+
\Delta^2\alpha_{|y|}.
\label{eq:pair_projection}
\end{equation}
A pairwise estimate that ignores triples therefore absorbs all third-order terms containing the pair $(i,j)$.

At a fixed rest configuration $y$, define
\begin{equation}
g_{ij}(y)=
\sum_{\ell\in\supp(y)}\theta_{ij\ell}.
\end{equation}
Under \cref{eq:pair_projection}, all third-order terms containing both $i$ and $j$ enter only the $11$ cell, through $x_i x_j g_{ij}(y)$.
If the true third-order coefficients are known, reweighting the four pair cells according to
\begin{equation}
\widetilde P_{ij,y}(a,b)
\propto
P(x_i=a,x_j=b,y)\exp\{-ab\,g_{ij}(y)\}
\label{eq:local_reweight}
\end{equation}
multiplies the $11$-cell mass by $\exp\{-g_{ij}(y)\}$ while leaving the other three cell masses unchanged before jointly normalizing all four cells.
Let $\widetilde\Lambda_{ij}(y)$ denote the order-$2$ contrast \cref{eq:fixedcontrast} for $T=\{i,j\}$, evaluated from the cell probabilities $\widetilde P_{ij,y}(a,b)$.
Then
\begin{equation}
\widetilde\Lambda_{ij}(y)
=
J_{ij}+\Delta^2\alpha_{|y|}.
\label{eq:local_removal}
\end{equation}
Thus the third-order contribution is removed exactly at fixed $y$.

The empirical analogue applies the same basket-level weighting with estimated coefficients and rest-cardinality aggregation.
Using CSID-3, define the estimated third-order contribution associated with pair $(i,j)$ in basket $x$ as
\begin{equation}
\widehat g_{ij}(x)=
\sum_{\ell:\{i,j,\ell\}\in\mathcal T_3}
\widehat\theta_{ij\ell}x_\ell.
\label{eq:gij}
\end{equation}
For $T=\{i,j\}$, $a,b\in\{0,1\}$, and rest cardinality $r$, we reconstruct the pairwise cell counts as
\begin{equation}
\widetilde n_{ij,r}^{ab}=
\sum_{n:\,x_i^{(n)}=a,\,x_j^{(n)}=b,\,R_T^{(n)}=r}
\exp\{-ab\,\widehat g_{ij}(x^{(n)})\}.
\label{eq:reweighted_counts}
\end{equation}
Only the $11$ cell is reweighted.
Each observed basket receives the same basket-level weight $\exp\{-ab\,\widehat g_{ij}(x^{(n)})\}$ as in \cref{eq:local_reweight}, but the sum aggregates over all rest configurations with common cardinality $r$ rather than over a single $y$.
Because baskets with the same $r$ can carry different weights when their rest configurations differ, $\widetilde n_{ij,r}^{ab}$ does not correspond to cell probabilities at any fixed $y$; it is a cardinality-stratified pseudo-count table.
Replacing the structural coefficients by CSID-3 estimates adds a second approximation.
Fitting CSID-2 to the reweighted counts yields a reweighted CSID-2 pairwise estimate.
The reweighting is local to each pair, does not define one common joint distribution, and is therefore an approximate pair-specific deprojection in empirical data rather than an exact recovery of a global pairwise model.

\section{Synthetic Validation}

We next evaluate CSID under controlled data-generating processes with known ground truth, before applying the method to public transaction data in the following section.
Unless stated otherwise, baskets were sampled independently from the exact categorical distribution implied by \cref{eq:model} on nonempty states.
All synthetic CSID fits used $\varepsilon=0.5$, $\lambda_2=\lambda_3=1$, and rest-cardinality strata $r\ge1$.

\subsection{Negative contamination of ordinary Ising couplings by cardinality structure}

We generated 10,000 baskets from a 10-item model with all true pairwise interactions fixed at $J_{ij}=0$.
For each replicate, fields were sampled as $h_i\sim N(-0.30,0.25^2)$ and the cardinality potential was
\begin{equation}
\alpha_m(s)=-s(m-1)^{1.3},\qquad m\ge1.
\end{equation}
We varied $s\in\{0,0.25,0.5,0.75,1.0\}$ and used 20 replicates per setting, holding $h_i$ fixed within each replicate so that only the cardinality potential changed.

Each sample was fitted with an ordinary pairwise Ising model without a cardinality term by exact maximum likelihood on the nonempty state space.
A ridge penalty of $10^{-5}$ was used for numerical stability.
CSID-2 was applied to the same sample with the settings reported above.
The penalties serve different numerical roles and are not calibrated to be directly comparable; this experiment tests directional contamination caused by omitting the cardinality term, rather than general predictive superiority of one estimator.

\begin{table}[t]
\centering
\small
\caption{
Ordinary Ising and CSID-2 estimates under a cardinality-only data-generating process.
Mean estimate is the Monte Carlo mean over 20 replicates of the within-replicate average across all $\binom{10}{2}=45$ fitted pair coefficients.
Neg.\ frac.\ and RMSE are averaged analogously from their within-replicate pair summaries; all true pair coefficients are zero.
}
\label{tab:cardinality}
\begin{tabular}{rrrrrrrr}
\toprule
$s$ & Mean size & \multicolumn{3}{c}{Ordinary Ising} & \multicolumn{3}{c}{CSID-2}\\
\cmidrule(lr){3-5}\cmidrule(lr){6-8}
& & Mean estimate & Neg. frac. & RMSE & Mean estimate & Neg. frac. & RMSE\\
\midrule
0.0 & 4.27 & -0.001 & 0.514 & 0.042 & 0.000 & 0.502 & 0.042\\
0.25 & 3.36 & -0.056 & 0.887 & 0.073 & -0.000 & 0.492 & 0.047\\
0.5 & 2.72 & -0.137 & 0.997 & 0.146 & -0.001 & 0.493 & 0.052\\
0.75 & 2.28 & -0.235 & 1.000 & 0.242 & -0.002 & 0.496 & 0.062\\
1.0 & 1.97 & -0.361 & 1.000 & 0.367 & -0.003 & 0.511 & 0.080\\
\bottomrule
\end{tabular}
\end{table}

\begin{figure}[t]
\centering
\includegraphics[width=0.88\linewidth]{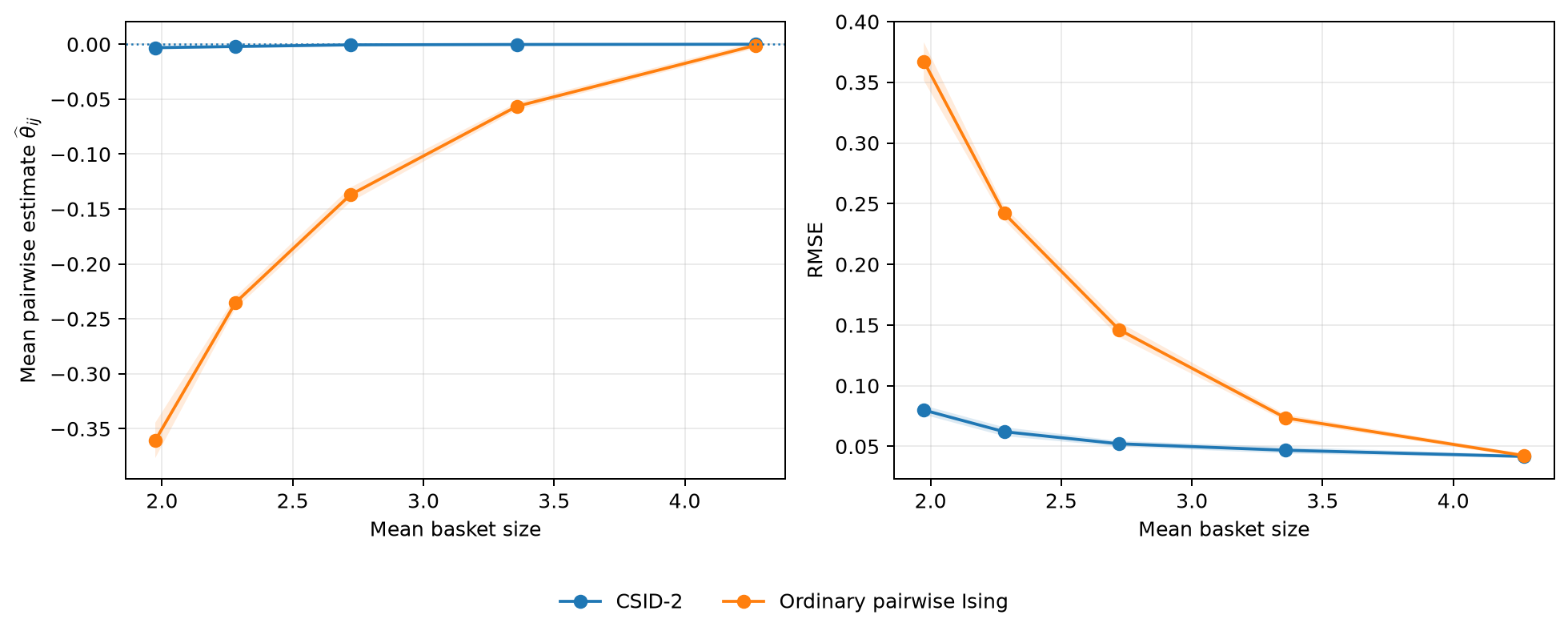}
\caption{
Mean pairwise estimate $\widehat{\theta}_{ij}$ (left) and RMSE to the true zero vector (right) against mean basket size.
Mean basket size increases from left to right; stronger cardinality penalties therefore appear farther left, and strength levels $s$ are listed in \cref{tab:cardinality}.
Lines compare ordinary pairwise Ising and CSID-2, with shaded bands showing 95\% Monte Carlo intervals over 20 replicates.
As baskets become smaller, ordinary Ising couplings shift downward, whereas CSID-2 remains near zero with substantially lower RMSE.
}
\label{fig:cardinality}
\end{figure}

\Cref{tab:cardinality,fig:cardinality} summarize the results.
When $s=0$, both methods were centered near zero.
As the cardinality constraint strengthened, ordinary Ising couplings shifted systematically downward; 99.7\% of fitted Ising pairs were negative at $s=0.5$, and every fitted pair was negative for $s\ge0.75$, whereas CSID-2 retained a sign split near 50\% negative and much smaller RMSE.
The optimizer reported convergence for 98 of the 100 ordinary-Ising fits; retaining the two flagged fits does not affect the directional pattern.
Thus an ordinary pairwise Ising model without a cardinality term can create apparently competitive edges even when no true pair-specific competition exists under this data-generating process.

\subsection{Detection of randomized sparse third-order signals}

For each replicate, we generated a new background with fields $h_i\sim N(-0.25,0.35^2)$, independently retained 35\% of pair coefficients with nonzero values drawn from $N(0,0.30^2)$, and sampled the cardinality-strength parameter uniformly from $[0.50,0.80]$.
We used ten items and 10,000 baskets.
These settings were chosen to generate heterogeneous but nondegenerate pairwise and basket-cardinality structure rather than to match a particular empirical dataset.
Eight of the $\binom{10}{3}=120$ triples were sampled uniformly, with four positive and four negative coefficients.
Absolute effect sizes were $0.1$, $0.2$, $0.3$, and $0.5$, with 20 replicates per effect size.
CSID-3 was fitted to all 120 triples, and screening performance was assessed with the score $z_T$ in \eqref{eq:zscore}.
We treated $|z_T|\ge2$ as a detection event; power and the non-planted exceedance rate count such events among planted and non-planted triples, respectively.
Although non-planted triples have zero generating third-order coefficients, their marginalized CSID contrasts need not be zero because marginalizing over the identities of rest items can transmit association from planted triples.
The non-planted exceedance rate therefore measures how often screening flags triples without a directly planted coefficient; it is not a type-I error rate under a global null of zero target contrasts.
We also report AUPRC, AUROC, and sign recovery, noting that precision-recall and ROC summaries respond differently to class imbalance \cite{davis2006relationship}.

\begin{table}[t]
\centering
\small
\caption{
Detection of randomized sparse third-order signals.
Non-planted rate is the fraction of non-planted triples with $|z_T|\ge2$.
Monte Carlo intervals are two-sided 95\% $t$ intervals over the 20 replicate-level detection rates.
}
\label{tab:power}
\begin{tabular}{rrrrrr}
\toprule
$|\theta|$ & Power (95\% MC interval) & Non-planted rate & AUPRC & AUROC & Sign recovery\\
\midrule
0.1 & 0.075 [0.035, 0.115] & 0.032 & 0.133 & 0.526 & 0.719\\
0.2 & 0.194 [0.142, 0.246] & 0.034 & 0.259 & 0.662 & 0.850\\
0.3 & 0.338 [0.235, 0.440] & 0.042 & 0.393 & 0.766 & 0.944\\
0.5 & 0.700 [0.589, 0.811] & 0.038 & 0.745 & 0.924 & 0.988\\
\bottomrule
\end{tabular}
\end{table}

\begin{figure}[t]
\centering
\includegraphics[width=0.88\linewidth]{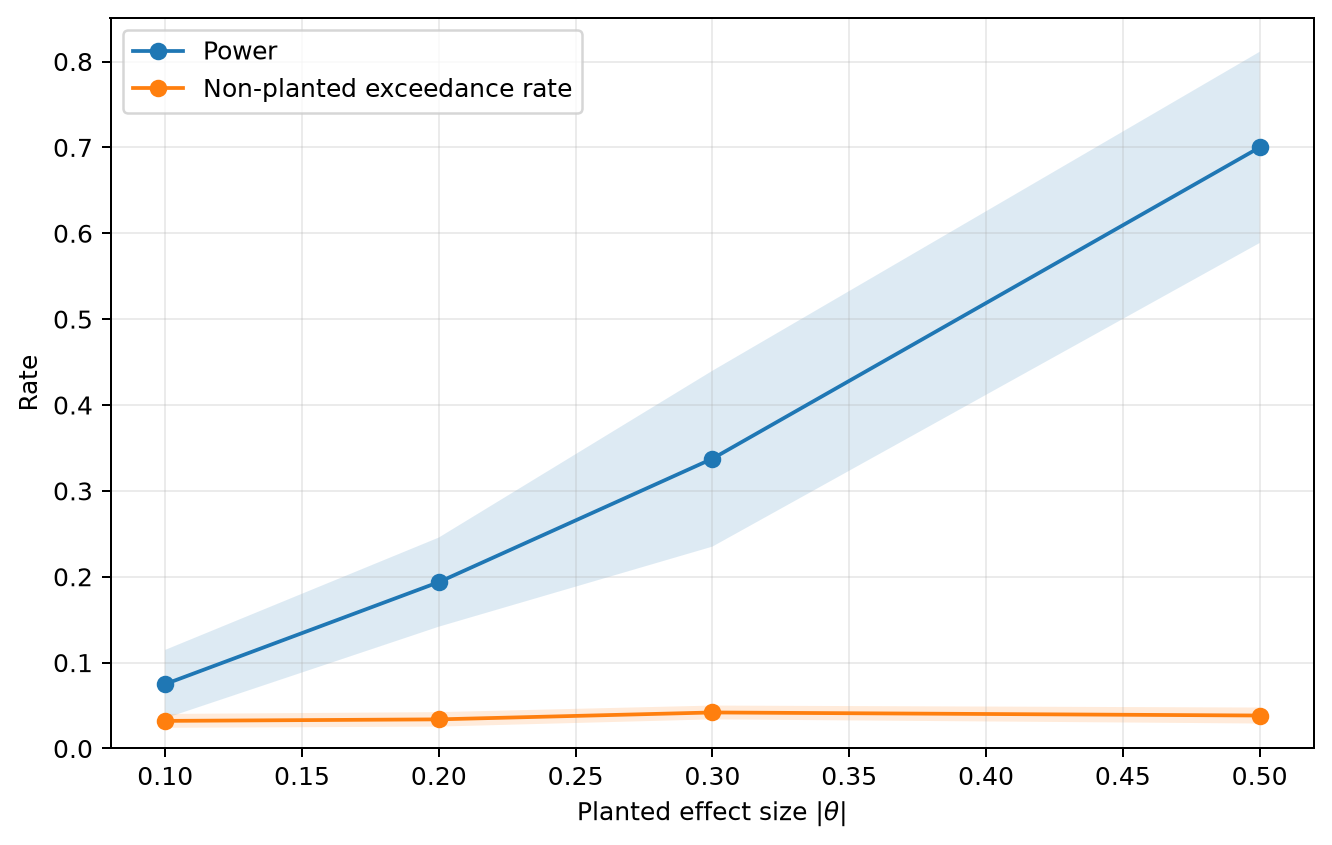}
\caption{
Planted-triple detection rate (power) and non-planted exceedance rate against planted effect size $|\theta|$.
The horizontal axis shows absolute planted coefficients; the vertical axis shows the fraction of fitted triples flagged.
Shaded bands are 95\% Monte Carlo intervals over 20 replicates (\cref{tab:power}).
Power increases with $|\theta|$, whereas the non-planted exceedance rate remains near 3--4\%.
}

\label{fig:power}
\end{figure}

\Cref{tab:power,fig:power} show that non-planted triples were rarely flagged and that this rate was stable across planted effect sizes.
Detection rose with $|\theta|$: weak planted coefficients were usually missed, whereas strong ones were detected substantially more often and typically with the correct sign.
The screening procedure is therefore conservative at this sample size, and is suitable for prioritizing candidates rather than exhaustively finding weak interactions.

\subsection{Recovery of pairwise structure under third-order saturation}

We sampled 60{,}000 baskets from a 6-item model with
\begin{equation}
J_{AB}=J_{AC}=J_{BC}=0.8,
\qquad
\theta_{ABC}=-1.2,
\label{eq:saturation}
\end{equation}
and background interactions $J_{DE}=0.5$ and $J_{EF}=0.3$, together with a cardinality potential.
Items $A,B,C$ are positively associated in pairs, but the negative triple coefficient induces saturation when all three are present.
A pairwise projection that omits the triple therefore underestimates the generating pair coefficients.

To evaluate recovery, we compared standard CSID-2 with reweighted CSID-2 from the local deprojection procedure (\cref{eq:gij,eq:reweighted_counts}).
For gauge alignment, the information-weighted mean of the generating pair vector, using the fitted CSID-2 weights $\mu_T$, was subtracted from every generating pair coefficient before RMSE and correlation were computed (\cref{eq:gauge}).

\begin{table}[t]
\centering
\caption{
Recovery of pair coefficients in one controlled 60{,}000-basket realization with third-order saturation.
RMSE is computed against the generating pair coefficients after information-weighted gauge alignment; Pearson $r$ is computed across all 15 item pairs.
}
\label{tab:saturation}
\begin{tabular}{lrr}
\toprule
Estimator & Gauge-aligned RMSE & Pearson $r$\\
\midrule
CSID-2 & 0.244 & 0.693\\
Reweighted CSID-2 & \textbf{0.073} & \textbf{0.985}\\
\bottomrule
\end{tabular}
\end{table}

\Cref{tab:saturation} shows that reweighted CSID-2 tracks the generating pair coefficients substantially more closely than CSID-2 alone.
In this controlled realization with known ground truth, the result demonstrates the underestimation expected when a triple is omitted from a pairwise projection (\cref{eq:pair_projection}).

\section{Evaluation on Public Transaction Data}

We next apply CSID to three public grocery transaction datasets in a matched cross-period design.
Synthetic validation established controlled behavior under known ground truth; the real-data evaluation asks whether CSID-2 and CSID-3 yield interpretable and reproducible structure when basket size, category popularity, and higher-order interaction are all present.

\subsection{Datasets and evaluation design}

Three public transaction datasets were each restricted to 20 selected categories for analysis (\Cref{tab:datasets}).
The Complete Journey data were obtained through the \texttt{completejourney} package \cite{boehmke2025completejourney}; transactions were joined to \texttt{products.csv}, and a predefined list of 20 \texttt{product\_category} values was retained from the full category catalog.
The Ta-Feng data were originally described by Hsu et al.\ \cite{hsu2004mining} and obtained from a Kaggle distribution \cite{tafengKaggle}; SKUs were aggregated to the first two digits of \texttt{PRODUCT\_SUBCLASS}, and the 20 codes with highest basket support over the full observation window were selected (minimum support 1{,}000).
The Instacart data came from the 2017 public release \cite{instacart2017dataset}, obtained from a Kaggle distribution \cite{instacartKaggle}; 30{,}000 users were sampled from prior orders, products were aggregated to aisles, and the 20 aisles with highest order support were selected.
In all cases, category presence was binarized within each basket and baskets with fewer than two selected categories were excluded.
Temporal splits assign weeks 1--26 to H1 and week 27 onward to H2 in Complete Journey, November--December 2000 to H1 and January--February 2001 to H2 in Ta-Feng, and a user's first ten orders to H1 and later orders to H2 in Instacart.
Category universes were selected before the temporal split and then held fixed in both periods.
All real-data fits used $\varepsilon=0.5$, $\lambda_2=\lambda_3=1$, and rest-cardinality strata $r\ge2$.
Candidates were enumerated exhaustively and filtered by the triple-count thresholds in \cref{tab:datasets}; the H1-retained candidate set was then fixed for H2 evaluation.
H2 and the full-period evaluation reference were fitted on this same candidate family using $g_T=\mu_T^{\mathrm{H1}}$ in \cref{eq:gauge}, placing all period-comparison coefficients in a common gauge while retaining each fit's own $\mu_T$ in \cref{eq:gamma,eq:theta_update}.

\begin{table}[t]
\centering
\footnotesize
\setlength{\tabcolsep}{4pt}
\caption{
Datasets and temporal splits after category restriction.
The column $n_{111}$ min.\ gives the minimum total triple co-occurrence count $\sum_r n_{111,r}$ required to retain a CSID-3 candidate in H1 (\cref{eq:cellcount}); pairs were retained without an analogous cutoff.
}
\label{tab:datasets}
\begin{tabularx}{\linewidth}{@{}l r c >{\raggedright\arraybackslash}X c@{}}
\toprule
Dataset & Baskets & H1 / H2 & Selected categories & $n_{111}$ min.\\
\midrule
Complete Journey & 72,936 & 35,087 / 37,849 & 20 predefined \texttt{product\_category} values & 0\\
Ta-Feng & 90,258 & 44,564 / 45,694 & Top 20 two-digit \texttt{PRODUCT\_SUBCLASS} codes & 10\\
Instacart & 373,133 & 179,784 / 193,349 & Top 20 aisles (30{,}000-user subsample) & 30\\
\bottomrule
\end{tabularx}
\end{table}

\Cref{tab:datasets} reports sample sizes, temporal splits, category definitions, and the $n_{111}$ minimum-count thresholds after this restriction.
With 20 selected categories per dataset, CSID enumerates 190 pair candidates and up to 1,140 triple candidates exhaustively.
Ta-Feng retains 1,066 triples in the full-sample fit and 1,001 H1-supported triples for H2 evaluation.

\subsection{Cross-period transfer of third-order effects}
\label{sec:cross-period-transfer}

We first fitted CSID-3 separately in H1 and H2, with the H2 candidate family and gauge fixed from H1 as described above.
Among common triple candidates, we ranked $\min(\mu_T^{\mathrm{H1}},\mu_T^{\mathrm{H2}})$ and evaluated the top information quartile using Pearson correlation, sign agreement, and overlap between the 20 most negative CSID-3 coefficients in H1 and H2.
Sensitivity to shared categories was assessed by removing, one at a time, all evaluated triples containing a category without refitting the models (leave-one-category-out evaluation, LOCO).

\begin{table}[t]
\centering
\footnotesize
\setlength{\tabcolsep}{4pt}
\caption{
H1--H2 reproducibility of high-information CSID-3 effects.
The LOCO range is the minimum and maximum Pearson correlation over the 20 omitted-category evaluations; the models are not refitted.
The top information quartile is reselected after each category omission.
The reported expectation is $k^2/N$, the expected overlap of two independently selected sets of $k=\min(20,N)$ triples among the $N$ evaluated triples.
}
\label{tab:stability}
\begin{tabularx}{\linewidth}{@{}l >{\centering\arraybackslash}X c >{\centering\arraybackslash}p{0.22\linewidth}@{}}
\toprule
Dataset & Pearson (LOCO range) & Sign agreement & Negative top-20 / expectation\\
\midrule
Complete Journey & 0.572 [0.539, 0.616] & 0.677 & 6 / 1.40\\
Ta-Feng & 0.739 [0.549, 0.768] & 0.729 & 11 / 1.59\\
Instacart & 0.723 [0.664, 0.742] & 0.747 & 9 / 1.40\\
\bottomrule
\end{tabularx}
\end{table}

\Cref{tab:stability} shows positive temporal correlation in all three datasets, directionally consistent descriptive LOCO ranges, and overlap among the 20 most negative third-order coefficients that exceeded random expectation.

We next assessed whether H1 effects reconstruct H2 stratified third-order contrasts.
For triple $T=\{i,j,\ell\}$ and rest-cardinality stratum $r$, we built the eight H2 cells and computed
\begin{equation}
\widehat\Lambda_{T,r}^{\mathrm{H2}}=
\sum_{A\subseteq T}(-1)^{3-|A|}
\log(n_{T,r}^{A,\mathrm{H2}}+\varepsilon).
\label{eq:h2_observed}
\end{equation}
Equivalently,
\begin{equation}
\widehat\Lambda_{ij\ell,r}^{\mathrm{H2}}=
\log\widehat{OR}_{ij\mid x_\ell=1,R_T=r}^{\mathrm{H2}}
-
\log\widehat{OR}_{ij\mid x_\ell=0,R_T=r}^{\mathrm{H2}}.
\label{eq:h2_or}
\end{equation}
This quantity is computed only from H2 cells; it contains neither H1 estimates nor fitted values from \cref{eq:objective}.

The projection \cref{eq:projection} motivates two reconstructions.
Without a triple-specific component, estimate the H2 component common to rest cardinality $r$ from all triples other than $T$,
\begin{equation}
\widehat\beta_{3,r}^{\mathrm{H2},(-T)}=
\frac{\sum_{U\neq T}w_{U,r}^{\mathrm{H2}}
\widehat\Lambda_{U,r}^{\mathrm{H2}}}
{\sum_{U\neq T}w_{U,r}^{\mathrm{H2}}},
\qquad
\widehat\Lambda_{T,r}^{\mathrm{base}}=
\widehat\beta_{3,r}^{\mathrm{H2},(-T)}.
\label{eq:baseline_reconstruction}
\end{equation}
For CSID-3 transfer, rewrite \cref{eq:projection} as
\begin{equation}
\beta_{3,r}\approx\Lambda_{U,r}-\theta_U.
\end{equation}
Remove the H1 triple-specific effect from the other H2 contrasts,
\begin{equation}
\widehat\beta_{3,r}^{\mathrm{H2}\mid\mathrm{H1},(-T)}=
\frac{\sum_{U\neq T}w_{U,r}^{\mathrm{H2}}
(\widehat\Lambda_{U,r}^{\mathrm{H2}}-
\widehat\theta_U^{\mathrm{H1}})}
{\sum_{U\neq T}w_{U,r}^{\mathrm{H2}}},
\label{eq:transferred_common}
\end{equation}
and add the H1 effect of the target triple,
\begin{equation}
\widehat\Lambda_{T,r}^{\mathrm{CSID3}}=
\widehat\beta_{3,r}^{\mathrm{H2}\mid\mathrm{H1},(-T)}
+
\widehat\theta_T^{\mathrm{H1}}.
\label{eq:csid_reconstruction}
\end{equation}
The subtraction in \cref{eq:transferred_common} extracts the H2 rest-cardinality-common component after removing other triples' H1-specific effects; the addition in \cref{eq:csid_reconstruction} transfers the target triple's H1-specific component into H2.
The target triple itself is excluded from estimation of the common component.
Both reconstructions therefore use H2 to estimate the common component.
The comparison is a partial-transfer evaluation of the incremental information in $\widehat\theta_T^{\mathrm{H1}}$, not a fully out-of-sample prediction of H2 contrasts from H1 alone.

As an external comparator, we fit a cardinality-aware higher-order pseudolikelihood model (CA-HOPL) to H1 only.
CA-HOPL represents each item's conditional log odds using incident pair and triple predictors together with a flexible intercept $a_{i,r}$ indexed by rest size $r=\sum_{j\ne i}x_j$.
Estimating these conditional models separately is the standard pseudolikelihood approach to binary graphical models, including models with higher-order interactions \cite{besag1975statistical,ravikumar2010high,hofling2009estimation,mukherjee2022estimation}:
\begin{equation}
\logit\Pr(X_i=1\mid X_{-i})=
a_{i,r}
+
\sum_{j:\{i,j\}\in\mathcal P}J_{ij}^{(i)}x_j
+
\sum_{\{j,\ell\}:\{i,j,\ell\}\in\mathcal T}
\theta_{ij\ell}^{(i)}x_jx_\ell.
\label{eq:ca_hopl}
\end{equation}
The pair and triple candidate sets $\mathcal P$ and $\mathcal T$ were exactly those generated from H1 for CSID.
The incident nodewise estimates were averaged for each pair and triple.
The symmetrized triple vector was then placed in the same H1-information-weighted reference gauge as CSID using the single global centering
\begin{equation}
\overline\theta_T^{\mathrm{PL}}
=\frac{1}{3}\sum_{i\in T}\widehat\theta_T^{(i)},
\qquad
\widetilde\theta_T^{\mathrm{PL}}
=
\overline\theta_T^{\mathrm{PL}}
-
\frac{\sum_{U\in\mathcal T}\mu_U^{\mathrm{H1}}\overline\theta_U^{\mathrm{PL}}}
{\sum_{U\in\mathcal T}\mu_U^{\mathrm{H1}}}.
\label{eq:ca_hopl_symmetrization}
\end{equation}
This operation subtracts the same constant from every symmetrized triple and therefore preserves all between-triple differences even when the candidate family is restricted.
We refer to this explicitly as a nodewise-symmetrized CA-HOPL comparator; it is not the optimizer of a joint pseudolikelihood with shared symmetric coefficients.
Ridge regularization was selected by H1-only 80/20 validation over $\{0.1,1,10,100,1000\}$, preserving customer groups where available; H2 was not used for tuning.
The centered shared H1 triple coefficients were substituted for $\widehat\theta_U^{\mathrm{H1}}$ in \cref{eq:transferred_common,eq:csid_reconstruction}, so CA-HOPL and CSID used identical H2 contrasts, leave-one-triple-out common components, layers, and evaluation weights.
The node-specific $a_{i,r}$ terms make CA-HOPL at least as flexible as a shared cardinality potential for this conditional comparison.

To quantify whether the reconstructions track H2 third-order contrasts in both direction and magnitude, we compared $\widehat\Lambda_{T,r}^{\mathrm{H2}}$ at each layer with three reconstructions: the common-effect baseline, CA-HOPL transfer, and CSID-3 transfer.
A layer is one common triple $T$ and rest-cardinality stratum $r$ with positive information in both periods.
Layer weights were the harmonic mean of H1 and H2 information,
\begin{equation}
\omega_{T,r}=
\frac{2w_{T,r}^{\mathrm{H1}}w_{T,r}^{\mathrm{H2}}}
{w_{T,r}^{\mathrm{H1}}+w_{T,r}^{\mathrm{H2}}}.
\label{eq:cross_weight}
\end{equation}
Collecting the selected layers into vectors $\widehat\Lambda^{\mathrm{H2}}$, $\widehat\Lambda^{\mathrm{base}}$, $\widehat\Lambda^{\mathrm{CA\text{-}HOPL}}$, and $\widehat\Lambda^{\mathrm{CSID3}}$, we define the information-weighted Lin concordance correlation coefficient (CCC) \cite{lawrence1989concordance} as
\begin{equation}
\CCC_{\omega}(u,v)=
\frac{2\Cov_{\omega}(u,v)}
{\Var_{\omega}(u)+\Var_{\omega}(v)
+(\bar u_{\omega}-\bar v_{\omega})^2}.
\label{eq:ccc}
\end{equation}
We evaluated
\begin{equation}
\begin{aligned}
\rho_{c,\omega}^{\mathrm{base}}
&=
\CCC_{\omega}(\widehat\Lambda^{\mathrm{H2}},
\widehat\Lambda^{\mathrm{base}}),
\\
\rho_{c,\omega}^{\mathrm{CA\text{-}HOPL}}
&=
\CCC_{\omega}(\widehat\Lambda^{\mathrm{H2}},
\widehat\Lambda^{\mathrm{CA\text{-}HOPL}}),
\\
\rho_{c,\omega}^{\mathrm{CSID3}}
&=
\CCC_{\omega}(\widehat\Lambda^{\mathrm{H2}},
\widehat\Lambda^{\mathrm{CSID3}}).
\end{aligned}
\label{eq:ccc_models}
\end{equation}
In \cref{eq:ccc_models}, each vector entry is one layer contrast and $\CCC_{\omega}$ uses $\omega_{T,r}$ from \cref{eq:cross_weight} as the stratum weight in the weighted mean, variance, and covariance across layers.
CCC decomposes as $\rho_{c,\omega}=\rho_{\omega}C_{b,\omega}$, where $\rho_{\omega}$ is the weighted Pearson correlation and $C_{b,\omega}$ is the bias-correction factor measuring mean and scale agreement.

\begin{figure}[t]
\centering
\includegraphics[width=0.96\linewidth]{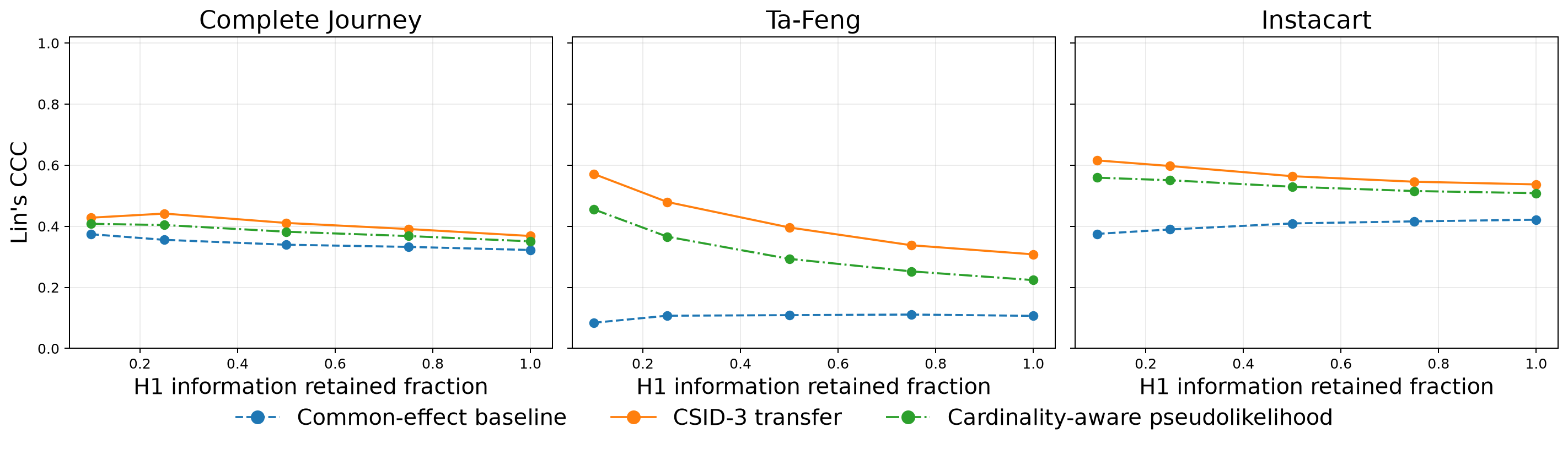}
\caption{
Information-weighted Lin's concordance correlation coefficient (CCC) between H2 stratified third-order contrasts and the common-effect baseline, CA-HOPL, and CSID-3 transfer reconstructions, plotted against the fraction of H2-evaluable triples retained by H1 information.
Each panel shows one dataset; all methods use identical triples, layers, H2 common-effect reconstruction, and information weights.
No H2 effect-size filter is applied.
CSID-3 CCC exceeds CA-HOPL at the primary 25\% fraction in every dataset.
}
\label{fig:ccc}
\end{figure}

To avoid selecting the comparison set with a CSID-specific score, the primary set retained the top 25\% of H2-evaluable triples by H1 information $\mu_T$.
No H2 effect-size threshold was imposed; every estimable layer of a retained triple entered \cref{eq:ccc_models} with weight \cref{eq:cross_weight}.
\Cref{fig:ccc} varies the retained H1-information fraction over 10\%, 25\%, 50\%, 75\%, and 100\%.
Primary top-quartile results are summarized in \Cref{tab:ccc,tab:ccc_decomp}; category leave-one-out (LOCO) sensitivity reselected the information quartile after each omission.

\begin{table}[t]
\centering
\footnotesize
\caption{
Continuous agreement with H2 third-order contrasts for the top H1-information quartile.
Baseline, CA-HOPL, and CSID-3 report information-weighted CCC on identical triples and layers; Gap is CSID-3 minus CA-HOPL.
The LOCO range is the minimum and maximum gap over the 20 omitted-category evaluations; the models are not refitted.
}
\label{tab:ccc}
\begin{tabular*}{\linewidth}{@{\extracolsep{\fill}}lrrrrrrr@{}}
\toprule
Dataset & Triples & Layers & Baseline & CA-HOPL & CSID-3 & Gap & LOCO range\\
\midrule
Complete Journey & 285 & 4,401 & 0.356 & 0.403 & \textbf{0.442} & 0.038 & [0.025, 0.049]\\
Ta-Feng & 251 & 2,472 & 0.107 & 0.358 & \textbf{0.480} & 0.122 & [0.083, 0.141]\\
Instacart & 285 & 3,812 & 0.390 & 0.549 & \textbf{0.597} & 0.049 & [0.039, 0.054]\\
\bottomrule
\end{tabular*}
\end{table}

\begin{table}[t]
\centering
\footnotesize
\caption{
Decomposition of weighted CCC for the common-effect baseline, CA-HOPL, and CSID-3 in the top H1-information quartile.
Following $\rho_{c,\omega}=\rho_\omega C_{b,\omega}$, $\rho_\omega$ is the weighted Pearson correlation between H2 contrasts and each reconstruction, and $C_{b,\omega}$ measures agreement in mean and scale.
}
\label{tab:ccc_decomp}
\begin{tabular*}{\linewidth}{@{\extracolsep{\fill}}lrrrrrr@{}}
\toprule
& \multicolumn{3}{c}{Weighted correlation $\rho_\omega$}
& \multicolumn{3}{c}{Bias correction $C_{b,\omega}$}\\
\cmidrule(lr){2-4}\cmidrule(lr){5-7}
Dataset & Baseline & CA-HOPL & CSID-3 & Baseline & CA-HOPL & CSID-3\\
\midrule
Complete Journey & 0.499 & 0.524 & 0.520 & 0.71 & 0.77 & 0.85\\
Ta-Feng & 0.225 & 0.454 & 0.512 & 0.48 & 0.79 & 0.94\\
Instacart & 0.495 & 0.619 & 0.637 & 0.79 & 0.89 & 0.94\\
\bottomrule
\end{tabular*}
\end{table}

\Cref{tab:ccc} shows that CSID-3 CCC exceeded both the common-component baseline and CA-HOPL in all datasets.
The CSID-3 minus CA-HOPL gap was 0.038 for Complete Journey, 0.122 for Ta-Feng, and 0.049 for Instacart; descriptive category-LOCO ranges were consistent in direction.
\Cref{tab:ccc_decomp} shows that CSID-3 had higher bias correction than CA-HOPL in all three datasets.
Its weighted correlation was also higher in Ta-Feng and Instacart, whereas CA-HOPL had a slightly higher correlation in Complete Journey; the Complete Journey CCC gain therefore arose from better mean and scale calibration.
Together, these correlation and calibration patterns show that the transfer comparison concerns continuous effect magnitude rather than sign alone.
\Cref{fig:ccc} shows the comparison as the H1-information restriction is varied without using method-specific scores.

\subsection{Full H1-only transfer}

As a secondary fully out-of-period diagnostic, we reconstruct H2 contrasts using H1 quantities only:
\begin{equation}
\widehat\Lambda_{T,r}^{\mathrm{full\ transfer}}
=
\widehat\beta_{3,r}^{\mathrm{H1}}
+
\widehat\theta_T^{\mathrm{H1}}.
\label{eq:full_transfer}
\end{equation}
Because \cref{eq:full_transfer} uses no H2 quantity, its evaluation also reflects any period shift in the common basket-cardinality environment.
Adding the H1 triple component improved CCC over the H1 common component alone in every dataset, although CCC remained below partial transfer (\cref{tab:full_transfer}).
Thus, conditional on the fixed H1-selected candidate family, the triple component adds out-of-period information.

\begin{table}[t]
\centering
\small
\caption{
Information-weighted CCC for H1-only reconstruction of H2 contrasts in the top H1-information quartile.
H1 common uses $\widehat\beta_{3,r}^{\mathrm{H1}}$ alone, Full CSID-3 uses \cref{eq:full_transfer}, Gain is their difference, and Partial CSID-3 repeats the primary reconstruction that estimates the common component from H2.
}
\label{tab:full_transfer}
\begin{tabular}{lrrrr}
\toprule
Dataset & H1 common & Full CSID-3 & Gain & Partial CSID-3\\
\midrule
Complete Journey & 0.032 & 0.158 & 0.127 & 0.442\\
Ta-Feng & -0.115 & 0.268 & 0.383 & 0.480\\
Instacart & 0.127 & 0.418 & 0.291 & 0.597\\
\bottomrule
\end{tabular}
\end{table}

\subsection{Conditional-odds interpretation}

To translate the global CCC result in \Cref{tab:ccc} into concrete item relations, we first restricted attention to triples with an H2-evaluable layer, retained the top H1 information quartile, and selected four negative and four positive triples by the H1 score $z_T$.
For each triple, the focal pair was the edge with the largest absolute H1 CSID-2 coefficient and the displayed rest-cardinality stratum was the H2-evaluable layer with greatest H1 information.
H2 therefore determines evaluability and supplies the observed contrast and row ordering, but is not used for sign-specific ranking.

\begin{figure}[t]
\centering
\includegraphics[width=\linewidth]{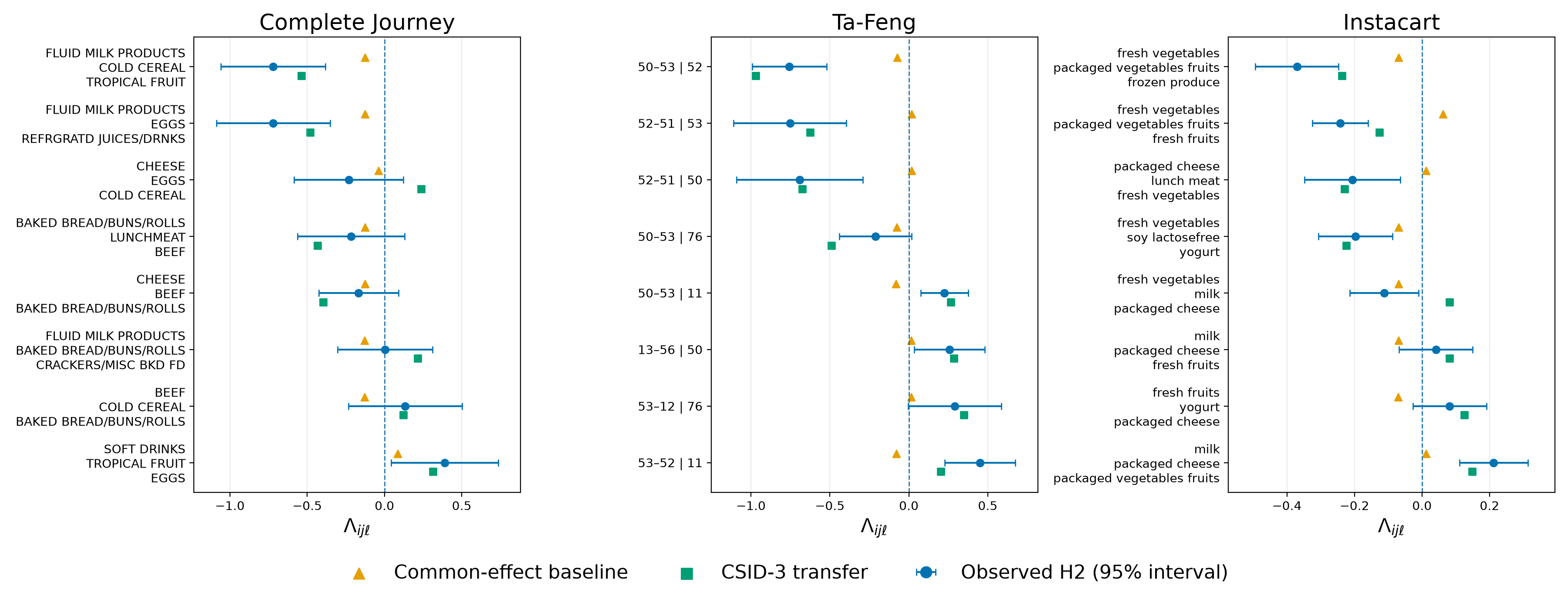}
\caption{
Conditional-odds contrasts $\Lambda_{ij\ell}=\log OR_{ij\mid x_\ell=1}-\log OR_{ij\mid x_\ell=0}$ for eight H1-ranked, H2-evaluable triples in each dataset.
Each panel shows one dataset; rows list the focal pair and moderator and are ordered by the observed H2 contrast.
Circles with horizontal intervals give observed H2 values with approximate 95\% cell-count delta-method reference intervals; these intervals do not account for customer clustering or selection of the displayed triples.
Triangles give the common-effect baseline and squares give the CSID-3 transfer.
A vertical dashed line marks $\Lambda=0$.
}
\label{fig:conditional}
\end{figure}

\Cref{fig:conditional} illustrates representative cases.
For Milk--Cereal moderated by Tropical fruit in Complete Journey, the H2 contrast was $\widehat\Lambda^{\mathrm{H2}}=-0.721$, indicating a weaker conditional pair odds ratio when Tropical fruit was present.
In Ta-Feng, the H2 contrast for pair 50--53 was negative with category 52 as moderator ($-0.757$) but positive with category 11 as moderator ($0.225$).
In Instacart, Fresh vegetables--Packaged vegetables/fruits had a negative H2 contrast when moderated by Fresh fruits ($-0.242$).

Across the displayed examples, transferring CSID-3 generally reproduced the direction and approximate magnitude of the observed modulation better than the common-component baseline, although over- and under-estimation remained for small effects.
\Cref{fig:conditional} is an interpretive visualization of the all-layer analysis in \Cref{tab:ccc,tab:ccc_decomp}, not an independent confirmatory test.

\subsection{Effect of third-order structure on pairwise projections}
\label{sec:pairwise-projection-effects}

Applying \cref{eq:gij,eq:reweighted_counts} and refitting CSID-2 changed pairwise projections in both directions.
\Cref{tab:pair_corrections} reports CSID-2, reweighted CSID-2, and their difference
$\Delta\theta^{(3)}:=\widehat\theta^{\mathrm{reweighted}}-\widehat\theta^{\mathrm{CSID2}}$.

\begin{table}[t]
\centering
\small
\caption{
Illustrative pairwise projections showing large positive and negative third-order corrections.
The rows were selected for interpretation rather than by a prespecified inferential rule.
}
\label{tab:pair_corrections}
\begin{tabularx}{\linewidth}{llrrr}
\toprule
Dataset & Pair & CSID-2 & Reweighted CSID-2 & $\Delta\theta^{(3)}$\\
\midrule
Complete Journey & Milk -- Cold cereal & 0.711 & 1.167 & 0.456\\
Complete Journey & Soup -- Crackers & 0.577 & 1.008 & 0.431\\
Ta-Feng & 51 -- 54 & 1.067 & 2.626 & 1.559\\
Ta-Feng & 73 -- 47 & -0.288 & -1.029 & -0.741\\
Instacart & Fresh herbs -- Cereal & -0.639 & -1.168 & -0.529\\
Instacart & Packaged cheese -- Lunch meat & 0.464 & 0.973 & 0.510\\
\bottomrule
\end{tabularx}
\end{table}

For Milk--Cold cereal and Ta-Feng pair 51--54, removing overlapping negative third-order effects strengthened the positive pairwise projection.
For Fresh herbs--Cereal, the negative pairwise projection became stronger, indicating that positive third-order components had partly offset it before deprojection.
\Cref{tab:pair_corrections} shows that the correction is therefore not a one-directional shrinkage reversal; it depends on the signs and locations of triples containing the pair.

To assess temporal stability, we applied H1 triple coefficients to H2 pair cells and compared the resulting correction with corrections using H2 coefficients and full-period coefficients.
\Cref{tab:pair_transfer} shows that H1-based corrections correlated positively with corrections estimated independently in H2 and were strongly correlated with the full-period reference.
The result indicates that the identity of the triples affecting each pair projection, and the direction of that effect, are not merely in-sample numerical adjustments.

\begin{table}[t]
\centering
\caption{
Cross-period reproducibility of third-order pairwise corrections.
Pearson correlations are computed across all 190 focal pairs between corrections obtained by applying H1 triple coefficients to H2 pair cells and corrections based on H2 or full-period fits.
All corrections use the same H1-fixed gauge convention, so the table compares relative correction patterns without changing the reference gauge.
}
\label{tab:pair_transfer}
\begin{tabular}{lrr}
\toprule
Dataset & Correlation with H2 & Correlation with full-period\\
\midrule
Complete Journey & 0.688 & 0.906\\
Ta-Feng & 0.695 & 0.904\\
Instacart & 0.814 & 0.945\\
\bottomrule
\end{tabular}
\end{table}

\subsection{Robustness to expansion of the item universe}

The main analyses used 20 selected categories to obtain a compact and interpretable item space in each dataset.
To examine whether the resulting interactions were artifacts of this truncation, we repeated the Instacart analysis while expanding the surrounding aisle universe from 20 to 40, 80, and all 134 aisles observed in the sampled users.

The focal top-20 aisles, the sampled users, the basket set, the H1--H2 split, and the focal pair and triple candidates were held fixed.
Only the aisles contributing to the rest-cardinality context were expanded.
All estimates were expressed in a common gauge defined by the information weights from the all-aisle analysis.

\begin{figure}[H]
\centering
\includegraphics[width=0.96\linewidth]{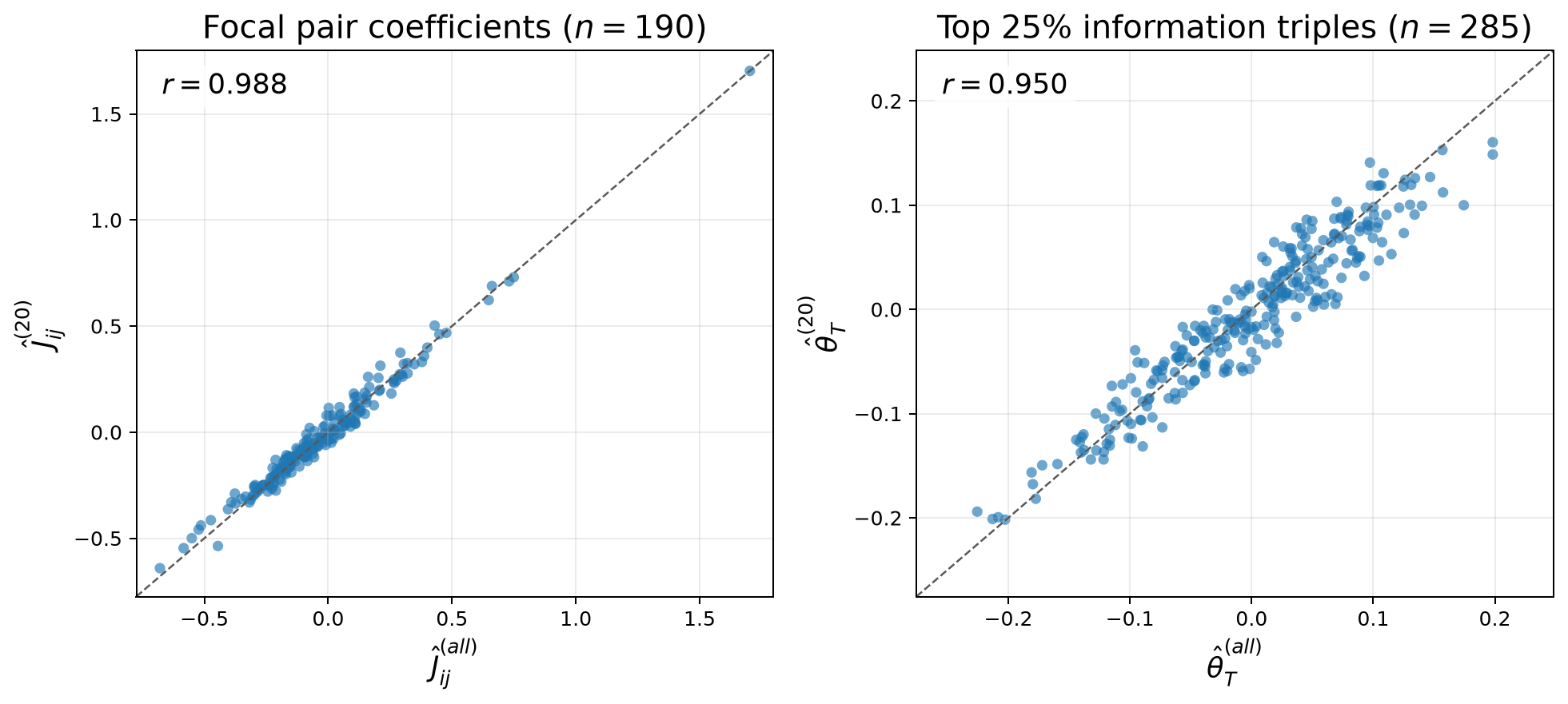}
\caption{
Coefficient stability under Instacart item-universe expansion.
Left: all 190 focal pair coefficients with 20 context aisles ($\widehat\theta_{ij}^{(20)}$) versus the all-aisle reference ($\widehat\theta_{ij}^{(\mathrm{all})}$).
Right: the top information quartile of focal third-order coefficients ($\widehat\theta_T^{(20)}$ versus $\widehat\theta_T^{(\mathrm{all})}$).
Dashed lines mark equality; annotations give Pearson correlation.
}
\label{fig:item_universe}
\end{figure}

As shown in \Cref{tab:item_universe,fig:item_universe}, focal pair and third-order coefficients remained highly correlated with the all-aisle reference at every context size, with Pearson correlations above 0.95 even when only 20 context aisles were used, and increased further as the universe expanded.
Third-order pairwise deprojection showed the same pattern.

These results indicate that the focal interaction structure obtained from the top-20 representation was not primarily induced by truncating the surrounding item universe.
The gain in cross-period CCC was nevertheless larger in the 20-aisle representation ($\Delta\mathrm{CCC}=0.286$) than in the expanded contexts ($0.163$--$0.172$; \Cref{tab:item_universe}).
Because expanding the item universe changes the definition of rest cardinality and hence the stratified contrast itself, we interpret this difference as context dependence of the transfer metric rather than instability of the focal coefficients.

\begin{table}[t]
\centering
\small
\caption{
Instacart robustness to expansion of the surrounding aisle universe.
Each row fixes the focal top-20 aisles and expands only the rest-cardinality context.
Pair $\rho$ compares all 190 focal $\widehat\theta_{ij}$ to the all-aisle reference; triple $\rho$ (top 25\%) compares the highest-information quartile of focal $\widehat\theta_T$; deprojection $\rho$ compares third-order pairwise corrections $\Delta\theta^{(3)}$.
All correlations use the all-aisle analysis as reference.
$\Delta\mathrm{CCC}$ is the CSID-3 minus baseline improvement in information-weighted Lin's CCC at $|z_T^{\mathrm{H1}}|\ge2$.
}
\label{tab:item_universe}
\begin{tabular}{lrrrr}
\toprule
Context aisles & Pair $\rho$ & Triple $\rho$ (top 25\%) & Deprojection $\rho$ & $\Delta\mathrm{CCC}$\\
\midrule
20 & 0.988 & 0.950 & 0.893 & 0.286\\
40 & 0.997 & 0.992 & 0.984 & 0.163\\
80 & 0.999 & 0.998 & 0.996 & 0.165\\
All (134) & 1.000 & 1.000 & 1.000 & 0.172\\
\bottomrule
\end{tabular}
\end{table}

The main temporal-reproducibility, partial-transfer, and pair-correction conclusions were stable across the tested smoothing constants, ridge penalties, and minimum rest-cardinality choices; detailed results are reported in Appendix~\ref{app:sensitivity}.

In a candidate-restricted audit on 373,133 Instacart baskets, the implementation retained 156,735 supported triple candidates across 134 aisles and completed the measured stages in 49.9 seconds; full implementation settings, runtime, and storage results are reported in Appendix~\ref{app:scaling}.

\section{Discussion}

\paragraph{Interpretive findings and empirical validation.}
The central result is that CSID separates item-set-specific association from a rest-cardinality-common environment and links the resulting pair and triple components to stratified log-odds contrasts with clear conditional-odds interpretations.
The temporal stability, transfer, and pairwise correction analyses provide empirical evidence that these components are not merely in-sample algebraic decompositions.
Specifically, the CSID-3 reconstruction combines the rest-cardinality-common component estimated from H2 with $\widehat\theta_T^{\mathrm{H1}}$ and yields higher weighted concordance with H2 contrasts than both the common-component baseline and CA-HOPL (\cref{eq:baseline_reconstruction,eq:csid_reconstruction}).
This is a partial-transfer rather than a fully out-of-sample prediction exercise.
The comparison to CA-HOPL is important because that model already contains pair, triple, and flexible rest-cardinality effects; the observed advantage therefore cannot be attributed simply to adding higher-order or cardinality terms.
The estimands nevertheless differ: CA-HOPL estimates symmetrized conditional-model coefficients, whereas CSID estimates components of marginalized contrast projections, so the benchmark compares their transfer to a common H2 contrast rather than equality of their coefficients.
Accordingly, the pseudolikelihood comparison evaluates agreement and transferability with respect to the stratified contrast targeted by CSID; it is not intended to establish universal predictive superiority.
A plausible explanation is target alignment: CSID directly estimates a projection of the same rest-cardinality-stratified marginalized contrasts used for evaluation, whereas CA-HOPL estimates conditional-model coefficients under an H1 conditional-log-loss objective.
The benchmark does not isolate this mechanism, but the CCC decomposition supports improved calibration as an important contributor to the observed advantage.
Together with the independently estimated H1--H2 coefficient stability and the conditional-odds examples, these results indicate that CSID-3 captures transferable information about both the direction and magnitude of third-item modulation of pair associations.
The H1-only diagnostic adds evidence that the transferred component is informative without using H2 in the reconstruction, while the sensitivity results in Appendix~\ref{app:sensitivity} support stability across the tested smoothing, ridge, and minimum-rest-size alternatives; partial transfer remains the primary estimand.

\paragraph{Interpretation and use.}
CSID-3 also provides a way to diagnose higher-order structure absorbed into pairwise projections (\cref{tab:pair_corrections,tab:pair_transfer,fig:conditional}).
The local reweighting analysis is useful for identifying pair projections whose apparent strength depends materially on overlapping triples.
Because the correction is pair-specific and does not define a common deprojected joint distribution, it should be regarded as a diagnostic decomposition rather than a replacement generative model.
In applications, the estimated signs and conditional-odds contrasts can prioritize item sets for substantive review, while the information score separates well-supported candidates from sparse ones.

\paragraph{Limitations.}
The fitted $\widehat\theta_T$ should be interpreted as an item-set-specific projection component of the marginalized contrasts, rather than necessarily as the natural parameter in \cref{eq:model}, and the estimated interactions are associative rather than causal.
In particular, the residual $\rho_{T,r}$ in \cref{eq:projection_residual} collects mediation and mixing induced by marginalizing the identities of rest items; the additive projection does not assume that this residual vanishes.
Likewise, the reported $z$ scores are information-normalized screening statistics, not calibrated independent-test $p$-values.
The real-data comparison estimates the common component from H2 and therefore establishes incremental transfer of the H1 triple component, not complete out-of-period prediction.
The Instacart item-universe analysis further showed that the focal pair and high-information triple coefficients were stable when the surrounding context was expanded, although the definition of the stratified contrast remains relative to the chosen item universe and candidate family.
The empirical local reweighting diagnostic also depends on the retained triple family, omitted interactions of order four and above, and the marginalization residual in \cref{eq:projection_residual}; it should not be interpreted as general recovery of structural pair coefficients.
The real-data analyses exclude baskets with fewer than two selected categories, so their conclusions concern multi-category baskets rather than the full transaction distribution that motivated the cardinality adjustment.
Finally, exhaustive candidate enumeration grows combinatorially with interaction order.
The scaling audit in Appendix~\ref{app:scaling} reached 134 aisles only through candidate restriction, so substantially larger or SKU-level item spaces still require prespecified, sparse, or distributed candidate generation.

\paragraph{Outlook.}
Further work could quantify the projection residual under broader data-generating processes, calibrate candidate screening under dependence, and compare CSID with coherent joint cardinality-aware graphical models on full-distribution predictive as well as interpretive criteria.
Extensions to customer segments or time-varying strata could also separate context-specific interaction changes from shifts in the overall basket-size environment.

\section{Conclusion}

CSID separates item-set-specific association from a rest-cardinality-common environment and makes explicit how omitted higher-order structure enters lower-order projections.
The theoretical contrast decomposition, controlled simulations, and three temporal grocery analyses---including a direct cardinality-aware higher-order pseudolikelihood comparison---support its use as an exploratory tool for signed pair and triple structure in transactional data.
By separating basket-cardinality structure from item-set-specific effects, CSID provides a practical basis for interpretable higher-order association analysis in sparse transactional data.

\appendix

\section{Parameter and design sensitivity}
\label{app:sensitivity}

For the analyses in \cref{sec:cross-period-transfer,sec:pairwise-projection-effects}, we tested the sensitivity of temporal $\rho$, the CSID--CA-HOPL partial-transfer gap, and the H1--H2 correlation of third-order-induced pairwise-projection changes.
\Cref{tab:sensitivity} reports default values and one-factor variations in $\varepsilon\in\{0.25,0.5,1\}$, $\lambda_2=\lambda_3\in\{0.1,1,10\}$, and inclusion of singleton-category baskets ($r\ge1$).
All three outcomes remained positive and close to their default values throughout.

\begin{table}[t]
\centering
\footnotesize
\caption{
Focused one-factor sensitivity analysis.
Temporal $\rho$ is the H1--H2 Pearson correlation in the top information quartile; CSID--CA-HOPL CCC is the partial-transfer gap; Correction $\rho$ compares H1-based and H2-based pair-correction patterns.
The $r\ge1$ analysis restores baskets containing one selected category before constructing the additional layer.
}
\label{tab:sensitivity}
\begin{tabular}{llrrrr}
\toprule
Dataset & Quantity & Default & $\varepsilon$ range & $\lambda$ range & $r\ge1$\\
\midrule
Complete Journey & Temporal $\rho$ & 0.572 & [0.571, 0.572] & [0.571, 0.572] & 0.647\\
 & CSID--CA-HOPL CCC & 0.038 & [0.038, 0.039] & [0.037, 0.038] & 0.049\\
 & Correction $\rho$ & 0.688 & [0.683, 0.699] & [0.688, 0.689] & 0.777\\
\addlinespace
Ta-Feng & Temporal $\rho$ & 0.739 & [0.736, 0.745] & [0.738, 0.745] & 0.779\\
 & CSID--CA-HOPL CCC & 0.122 & [0.121, 0.122] & [0.114, 0.123] & 0.123\\
 & Correction $\rho$ & 0.695 & [0.681, 0.721] & [0.687, 0.746] & 0.763\\
\addlinespace
Instacart & Temporal $\rho$ & 0.723 & [0.723, 0.723] & [0.723, 0.723] & 0.757\\
 & CSID--CA-HOPL CCC & 0.049 & [0.048, 0.049] & [0.048, 0.049] & 0.046\\
 & Correction $\rho$ & 0.814 & [0.814, 0.815] & [0.814, 0.815] & 0.859\\
\bottomrule
\end{tabular}
\end{table}

\section{Scaling audit}
\label{app:scaling}

To separate item-universe robustness from computational scaling, we reran candidate generation and estimation on the same 373,133 Instacart baskets while increasing the aisle universe (\cref{tab:scaling}).
At 20 and 40 aisles, sparse generation retained every supported exhaustive triple.
At 134 aisles it generated the configured cap of 250,000 candidates, retained 156,735 after the support filter, and completed the measured stages in 49.9 seconds using 178 MB of stratified array storage.
This demonstrates candidate-restricted scaling at the aisle level, not exhaustive or SKU-level scalability.

\begin{table}[t]
\centering
\small
\caption{
Single-process candidate-scaling audit on 373,133 Instacart baskets, conducted on a laptop computer with an arm64 processor.
Time is the sum of basket scanning, candidate construction, exact counting, contrast construction, and CSID-2/3 fitting; storage is the combined size of the retained stratified count and contrast arrays, not peak process memory.
Sparse generation used frequent-pair triangles and bounded neighborhoods with a 250,000-candidate cap.
}
\label{tab:scaling}
\begin{tabular}{rlrrrr}
\toprule
Aisles & Mode & Generated & Retained & Time (s) & Storage (MB)\\
\midrule
20 & Exhaustive & 1,140 & 1,140 & 6.9 & 0.6\\
20 & Sparse & 1,140 & 1,140 & 6.9 & 0.6\\
40 & Exhaustive & 9,880 & 9,880 & 16.5 & 6.8\\
40 & Sparse & 9,880 & 9,880 & 16.5 & 6.8\\
80 & Sparse & 82,160 & 79,193 & 37.6 & 83.0\\
134 & Sparse & 250,000 & 156,735 & 49.9 & 178.2\\
\bottomrule
\end{tabular}
\end{table}

\section*{Code availability}

Source code and scripts for reproducing the analyses and figures are available at \url{https://github.com/kavvase/csid}.

\bibliographystyle{JHEP}
\bibliography{references}

\end{document}